\documentclass{templates/mva2025/mva_style}

\usepackage{graphicx}
\usepackage{color}
\usepackage{amsmath}
\usepackage{amssymb}
\usepackage{multirow}
\usepackage{makecell}

\finalcopy 

\usepackage{makecell}

\begin{document}
\title{Incremental dimension reduction for efficient and accurate visual anomaly detection} 

\author{
	Teng-Yok Lee\\
	Information R\&D Technology Center\\
	Mitsubishi Electric Coporation\\
	{\tt Lee.Teng-Yok@ap.MitsubishiElectric.co.jp}\\
}

\maketitle

\begin{abstract}

While nowadays visual anomaly detection algorithms use deep neural networks to extract salient features from images, the high dimensionality of extracted features makes it difficult to apply those algorithms to large data with 1000s of images. To address this issue, we present an incremental dimension reduction algorithm to reduce the extracted features. While our algorithm essentially computes truncated singular value decomposition of these features, other than processing all vectors at once, our algorithm groups the vectors into batches. At each batch, our algorithm updates the truncated singular values and vectors that represent all visited vectors, and reduces each batch by its own singular values and vectors so they can be stored in the memory with low overhead. After processing all batches, we re-transform these batch-wise singular vectors to the space spanned by the singular vectors of all features. We show that our algorithm can accelerate the training of state-of-the-art anomaly detection algorithm with close accuracy.

\end{abstract}

\section{Introduction}

Based on deep neural networks, which can be pre-trained on large image datasets like ImageNet~\cite{:/journal/ijvc/2015/russakovsky_imagenet}, visual anomaly detection task can use these neural networks to extract salient features of images. Recent methods~\cite{:/misc/arxiv/2020/defard_padim,:/misc/arxiv/2020/rippel_mahalanobis_ad,:/misc/arxiv/2021/roth_patchcore,ieeexplore:/conf/icip/2022/ndiour_subspace} divide image domains into patches, and use deep neural networks like WideResNet50~\cite{:/conf/bmvc/2016/zagoruyko_wrn} to convert each image patch into a feature vector. Given a set of training images, these methods extract patch-wise feature vectors of all training images. When testing a new image, these methods also divide this image into patches, and compare the feature of each patch against the training features. While most methods compare features of a patch against features of the same patch, PatchCore~\cite{:/misc/arxiv/2021/roth_patchcore} compare against features of all image patches. Because all patches are considered, PatchCore can detect anomaly of moved and rotated objects more accurately than other methods. 

On the other hand, storing the feature vectors of all patches can be storage-consuming, and the comparison between vectors can be slow. Therefore, PatchCore samples a subset of feature vectors to reduce the overhead. In order to achieve high accuracy, PatchCore recommends using 10\% to 25\% of all features, which can be still large. Besides, the sampling needs to compute the distances between all pairs of features, which can be slow. The sampling speed is also related to the dimensionality of feature vectors, as the time complexity to compare a pair of $m$-dimensional vectors is $O(m)$. As the extracted features can have hundreds to thousands of dimensions, the dimensionality can further slow down the sampling. The testing speed could be impacted too, because it needs to compare the features of all patches against the sampled ones. Namely, the size of memory bank makes it difficult to apply PatchCore to large data sets with 1000s of images or more. 

To make PatchCore practical for large data sets, the speed overhead should be reduced. A straight forward approach is using dimension reduction techniques like singular value decomposition (SVD) to reduce the feature vectors. Based on SVD, we can simultaneously obtain the reduced feature vectors in lower dimensional space, and the corresponding basis vectors of the space. When testing a new feature vector, we can project this testing vector to this reduced space and compare the projected one against the training vectors. 

Conventional SVD algorithms, nevertheless, assume that all feature vectors are stored in the memory beforehand, which could be challenging for data sets with 1000s to 10000s of images. While there are online methods such as incremental SVD~\cite{:/journal/laa/2006/brand_incremental_svd} to reduce the dimensionality, those algorithms can get slower when processing later vectors. This is because that after using a new set of vectors to update the basis vectors, these methods will re-transform all visited vectors, not just the latest ones. Another type of online algorithms might only update the basis vectors, such as incremental principal component analysis (PCA) for object tracking~\cite{:/conf/nips/2024/lim_incemental_pca_visual_tracking}, but those algorithms need extra passes to reduce the feature vectors, which involve extra I/O and computation time to re-extract the features.

In order to efficiently utilize dimension reduction for PatchCore, in this paper, we present a new incremental algorithm that combines the ideas of of Incremental SVD~\cite{:/journal/laa/2006/brand_incremental_svd} and Incremental PCA~\cite{:/conf/nips/2024/lim_incemental_pca_visual_tracking}. Similar to these methods, our algorithm groups the vectors into batches. When processing each batch, our algorithm updates the singular values and singular vectors that correspond to the basis of reduced dimensions. To avoid extra passes to revisit the batches, our algorithm applies truncated SVD to reduce each batch so they can be stored in the memory. Once processing all batches, our algorithm reconstructs each batch based on its own singular values and vectors, and re-transforms the reconstructed batch to the space spanned by the singular vectors of all vectors. With our algorithm, we can accelerate the training speed, reduce the memory usage, and preserve the accuracy for visual anomaly detection. We tested PatchCore with out algorithm on a large data set with 1000s of images, which is difficult to achieve without dimension reduction.

\begin{figure}
	\centering
	\includegraphics[width=\linewidth]{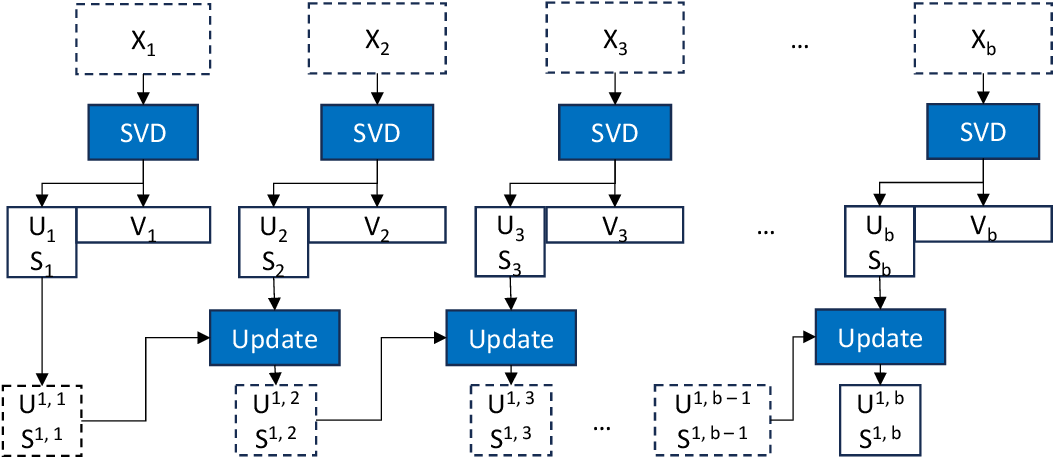} 
	\caption{
Overview of the algorithm that use a batch of vectors $X_b$ to update the singular values $S^{1,b}$ and vectors $U^{1,b}$. The variables in the dashed boxes will be released after the corresponding operation.
	}
	\label{fig:algorithm_update}
\end{figure}

\begin{figure}
	\centering
	\includegraphics[width=\linewidth]{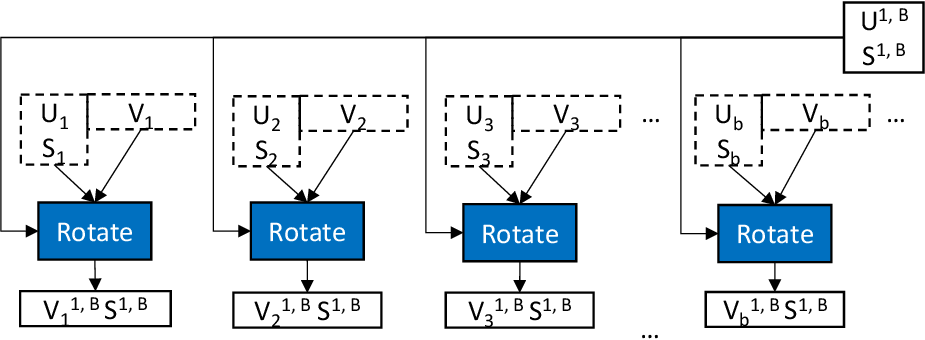} \\
	\caption{
Rotation of all projected $V_b$ to the final space of $U^{1,B}$. The variables in the dashed boxes will be released after the corresponding operation.
	}
	\label{fig:algorithm_finalize}
\end{figure}

\section{Incremental dimension reduction}

This section describes our dimension reduction algorithm. Given a training set of images, our algorithm considers all of their feature vectors as a matrix $X$ where each column represents one feature vector. Given $n$ feature vectors of $m$ dimensions, the size of $X$ is $m \times n$. Our goal is computing the truncated SVD of $X$ such that $X \approx U S V^{\intercal}$ where S is a diagonal matrix with the $k$ largest singular values $s[1], \dots, s[k]$ along its diagonal, and the columns of $U$ and $V$ are the corresponding singular vectors. The columns of matrix $V^{\intercal}$ represent the reduced features, and the columns of $U$ represent the basis of their corresponding space. When testing a new feature vector $x'$, we transform $x'$ to the same space of the reduced features by multiplying $x'$ by $\hat{U}^{\intercal}$. 

Because it could be memory-consuming to store the entire $X$ in the memory, we group column vectors of $X$ into batches, and then iterate through batches to update. Figure~\ref{fig:algorithm_update} overviews the iteration algorithm. For each batch, says $X_b$, our algorithm first computes its truncated SVD such that $X_b \approx U_b S_b V_b^{\intercal}$, and then updates the singular vectors and values for batches $X_1, \dots X_b$. Hereafter we use $S^{1, b}$ and $V^{1, b}$, respectively, to denote the updated singular values and vectors; namely, the superscript indicates the range of batches that have been iterated. 

Given $S_b$ and $V_b$ of the current batch, our algorithm updates $S^{1, b - 1}$ and $V^{1, b - 1}$ to $S^{1, b}$ and $V^{1, b}$. The idea behind the update is modifying the Gram matrix of $\left [X_1, \dots, X_b \right ]$, the concatenation of all visited batches $X_1 \dots X_b$, as shown in Equation~\ref{equ:gram_matrix_update}:

\begin{eqnarray}
	& & \left [X_1, \dots, X_b \right ] \left [X_1, \dots, X_b \right ]^{\intercal} 
	\nonumber
	\\
	& = &
	\left [X_1, \dots, X_{b - 1} \right ] \left [X_1, \dots, X_{b - 1} \right ]^{\intercal} + X_b X_b^{\intercal} 
	\label{equ:gram_matrix_update}
	\end{eqnarray}

\noindent With the largest truncated singular values $S^{1,b-1}$ and $U^{1, b-1}$, we can approximate the update of Gram matrix by Equation~\ref{equ:incremental_pca}:

\begin{eqnarray}
& 	& \left [X_1, \dots, X_{b - 1} \right ] \left [X_1, \dots, X_{b - 1} \right ]^{\intercal} + X_b X_b^{\intercal} 
\nonumber
\\
& \approx & \left [ U^{1, b-1} S^{1, b-1}, X_b \right] \left [ U^{1, b-1} S^{1, b-1}, X_b \right]^{\intercal}
\label{equ:incremental_pca}
\end{eqnarray}

In other word, by computing the truncated SVD of $[U^{1, b-1} S^{1, b-1}, X_b]$, the singular values and left singular vectors become the new $S^{1, b}$ and $U^{1, b}$. It should be noted that this approach is similar to the Incremental PCA algorithm presented by Lim \textit{et al.}~\cite{:/conf/nips/2024/lim_incemental_pca_visual_tracking}. In their algorithm, the goal is incrementally updating the SVD of a covariance matrix, which can be transformed from a Gram matrix after the removal of mean from all columns and the cancellation of bias. Because our interest is the Gram matrix itself, not the covariance matrix, there is no need to consider mean and bias here. 

After iterating batch $X_b$, our algorithm discards previous $U^{1, b-1}$ and $S^{1, b-1}$, and keeps its SVD matrices $U_b$, $S_b$ and $V_b$ in the memory. Once all batches are iterated, our algorithm transforms the reduced vectors $V_b$ of each batch to the space spanned by $U^{1, B}$, as shown in Figure~\ref{fig:algorithm_finalize}. With the reduced vectors of all batches in the same space, PatchCore can be applied to compute distances among them and sample the subset. 

This is the difference between our algorithm and existing incremental approaches~\cite{:/journal/laa/2006/brand_incremental_svd}, which are designed for scenario when $V$ of all visited batches are immediately needed after each iteration. When processing a new batch $X_b$, once $U^{1, b}$ and $S^{1, b}$ are updated, these algorithms re-transform all previous batches $V_1, \dots, V_{b-1}$ based on the updated $U^{1, b}$ and $S^{1, b}$. Consequently, larger the $b$ is, slower the update becomes.


\begin{table*}[tb!]
	\centering
	\caption{
		Average AUROC on MVTec AD~\cite{:/journal/ijvc/2021/bergmann_mvtec}. Different numbers ($k$) of singular values and batch sizes ($n_b$) were tested when using dimension reduction.
	}
	\label{table:mvtec_ad}
	\setlength\tabcolsep{4pt} 
	\begin{tabular}{lcc|c|ccc|ccc|c|ccc|ccc}
		&  
		& 
		& \multicolumn{7}{c}{WideResNet50~\cite{:/conf/bmvc/2016/zagoruyko_wrn}} 
		& \multicolumn{7}{|c}{ResNet18~\cite{:/conf/cvpr/2016/he_resnet}} 
		\\
		\hline
		& \multirow{2}{*}{\makecell{Update \\ parameter}}
		& $k$ 
		& N/A 
		& \multicolumn{3}{c|}{128} 
		& \multicolumn{3}{c|}{256}  
		& N/A 
		& \multicolumn{3}{c|}{128} 
		& \multicolumn{3}{c}{256}  
		\\
		& 
		& $n_b$ 
		& N/A 
		& 16K & 32K & All 
		& 16K & 32K & All  
		& N/A 
		& 16K & 32K & All 
		& 16K & 32K & All  
		\\
		\hline
		& \multirow{3}{*}{\makecell{Image \\ AUROC}} & All 
		& 99.1 & 98.9 & 98.9 & 98.9 & 99.1 & 99.1 & 99.0 & 97.3 & 96.9 & 97.1 & 96.9 & 97.2 & 97.2 & 97.1 \\
		& & Texture 
		& 99.3 & 98.9 & 99.1 & 99.0 & 99.3 & 99.2 & 99.2 & 97.9 & 96.7 & 96.9 & 96.3 & 97.5 & 97.3 & 97.1 \\
		& & Object 
		& 99.0 & 98.9 & 98.9 & 98.9 & 99.0 & 99.0 & 99.0 & 97.0 & 97.1 & 97.2 & 97.1 & 97.0 & 97.2 & 97.1 \\
		\hline 
		& \multirow{3}{*}{\makecell{Pixel \\ AUTOC}} & All 
		& 97.9 & 97.7 & 97.7 & 97.7 & 97.8 & 97.8 & 97.8 & 96.9 & 96.6 & 96.6 & 96.5 & 96.8 & 96.8 & 96.8 \\
		& & Texture 
		& 97.3 & 97.2 & 97.2 & 97.2 & 97.3 & 97.3 & 97.3 & 96.1 & 95.4 & 95.4 & 95.4 & 95.9 & 95.9 & 95.9 \\
		& & Object 
		& 98.2 & 98.0 & 98.0 & 98.0 & 98.1 & 98.1 & 98.1 & 97.3 & 97.2 & 97.1 & 97.1 & 97.3 & 97.3 & 97.3 \\
	\end{tabular}		
\end{table*}



Our final transformation, conceptually, first reconstructs each batch $X_b$ by its singular values/vectors, and then projects the reconstructed $X_b$ to the space spanned by the final $U^{1, b}$ and $S^{1, b}$, as shown in Equation~\ref{equ:batch_wise_projection}:

\begin{equation}
(V_b^{1,B})^{\intercal} ~\approx (S^{1,B})^{-1}(U^{1, B})^{\intercal} U_b S_b V_b^{\intercal} 
\label{equ:batch_wise_projection}
\end{equation}

When using GPUs, there are two issues to consider. First, if we first scale vectors by $S_b$ and then re-scale by the inverse of $S^{1,B}$, it could be numerically unstable with single-precision. Second, as the reconstructed $X_b'$ requires $O(m \times n_b)$ to store, if we reconstruct all batches in parallel, we need to restore the entire $X$ in the memory, which can be large. Based on both issues, the transform first compute a batch-wise rotation matrix $R_b$ by Equation~\ref{equ:batch_wise_rotation}:

\begin{equation}
R_b=(U^{1,B})^{\intercal} U_b S_b
\label{equ:batch_wise_rotation}
\end{equation}

This matrix $R_b$ essentially scales and rotates the $k$-dimensional vectors of $V_b$ from the space of $U_b$ to that of $U^{1,B}$. Because $R_b$ is a $k \times k$ matrix, it is small to store, and can be efficiently computed by $O(k \times m \times k)$ multiplications. Besides, $R$ will not scale the rotated vectors again.

After transforming $V_b$ of all batches, we follow conventional PatchCore algorithm~\cite{:/misc/arxiv/2021/roth_patchcore} for both training and testing. The training algorithm samples representative vectors from the reduced ones to form a code book called \emph{memory bank}. Given a new vector $y$ to examine, the testing algorithm transforms $y$ to $(U^{1,B})^{\intercal} y$, and computes the anomaly score of $y$ as the distance from $(U^{1,B})^{\intercal} y$ to the nearest entry in the memory bank.

\section{Experiments}
\label{sec:experiment}

To evaluate our incremental dimension reduction, we re-implemented and extended PatchCore~\cite{:/misc/arxiv/2021/roth_patchcore} with our algorithm. Our implementation used python 3, pytorch 1.12 and cuda 11.3, and tested on a Ubuntu 20 computer with Intel Xeon Gold 6230 CPU (2.10GHz), one nVidia QuadroRTX8000 GPU, and 11.0GB of RAM. While we followed the same approach as the original implementation of PatchCore to divide images and extract patch-wise features, we revised the sampling scheme to avoid randomness. Before sampling vectors into the memory bank, original PatchCore implementation randomly selects a small subset called \emph{anchors}. From the non-anchored vectors, their sampling scheme finds the farthest one from the anchors, adds this one to the memory bank and anchors, and repeats till the size of memory bank reaches user-specified limit. Because the initial selection of anchors can introduce randomness, we use the average of all vectors as the initial anchor, which is determined. Our experiments show that our implementation can achieve similar accuracy as the original PatchCore. 

We first show the results of WideResNet50~\cite{:/conf/bmvc/2016/zagoruyko_wrn} and ResNet18~\cite{:/conf/cvpr/2016/he_resnet} as the backbones. We resized each input image into $256 \times 256$ pixels, and cropped the central $224 \times 224$ pixels as the input. We use feature maps of the second and third layers, which essentially divide the image space into $28 \times 28$ patches. For each patch, PatchCore extracts its feature vectors by first re-sampling the corresponding feature vectors of both layers into 1024 dimensions, concatenating the resamples features of its $3 \times 3$ neighbors, and further re-sampling the concatenated one into 1024 dimensions. With WideResNet50 and this feature extraction scheme, PatchCore achieves state-of-the-art accuracy on dataset MVTec AD~\cite{:/journal/ijvc/2021/bergmann_mvtec}. 

Table~\ref{table:mvtec_ad} shows the image-wise and pixel-wise AUROC on MVTec AD. We tested different numbers of singular values ($k$) and batch sizes ($n_b$) to update, whereas $n_b=\text{all}$ means applying conventional SVD to all vectors once. The columns that are marked by N/A for both $k$ and $n_b$ show the results without dimension reduction. For WideResNet50, the column of $k=128$ and $n_b=16K$ of Tables~\ref{table:mvtec_ad} shows slightly lower image-level AUROC (98.9\%) than those of the original PatchCore (99.0\%), but  pixel-level AUROC (97.9\%) are close. 

\begin{table}[tb!]
	\centering
	\caption{
Category-wise Image AUROC (\%) on MVTec AD~\cite{:/journal/ijvc/2021/bergmann_mvtec} with WideResNet50~\cite{:/conf/bmvc/2016/zagoruyko_wrn}. Different numbers ($k$) of singular values and batch sizes ($n_b$) were tested when using dimension reduction.
	}
	\label{table:mvtec_ad_wide_resnet50}
	\setlength\tabcolsep{2pt} 
	\begin{tabular}{ll|r|rrr|rrr}
& $k$ 
& N/A 
& \multicolumn{3}{|c}{128} 
& \multicolumn{3}{|c}{256}  
		\\
		& $n_b$ 
		& N/A 
		& 16K & 32K & All 
		& 16K & 32K & All  
		\\
		\hline
& carpet & 99.3 & 98.0 & 98.1 & 98.1 & 98.7 & 98.8 & 98.9 \\
& grid & 99.3 & 98.9 & 99.5 & 99.0 & 99.5 & 99.1 & 99.2 \\
& leather & 100.0 & 100.0 & 100.0 & 100.0 & 100.0 & 100.0 & 100.0 \\
& tile & 99.3 & 99.3 & 99.5 & 99.4 & 99.3 & 99.2 & 99.3 \\
& wood & 98.9 & 98.4 & 98.3 & 98.5 & 98.8 & 98.8 & 98.8 \\
\hline
& bottl & 100.0 & 100.0 & 100.0 & 100.0 & 100.0 & 100.0 & 100.0 \\
& cable & 99.6 & 100.0 & 99.9 & 100.0 & 99.8 & 99.8 & 99.8 \\
& capsule & 96.7 & 96.7 & 96.6 & 96.3 & 97.4 & 97.0 & 97.1 \\
& hazelnut & 100.0 & 100.0 & 100.0 & 100.0 & 100.0 & 100.0 & 100.0 \\
& metal nut & 100.0 & 99.9 & 99.9 & 99.8 & 99.8 & 99.9 & 99.9 \\
& pill & 95.7 & 95.2 & 95.0 & 95.5 & 95.4 & 95.7 & 94.8 \\
& screw & 98.7 & 97.9 & 97.8 & 97.8 & 97.9 & 98.5 & 98.4 \\
& toothbrush & 100.0 & 99.7 & 100.0 & 100.0 & 100.0 & 100.0 & 100.0 \\
& transistor & 100.0 & 100.0 & 100.0 & 100.0 & 100.0 & 100.0 & 100.0 \\
& zipper & 99.4 & 99.6 & 99.6 & 99.6 & 99.6 & 99.6 & 99.6 \\
	\end{tabular}		
\end{table}



Table~\ref{table:mvtec_ad_wide_resnet50} lists the category-wise accuracies on MVTec AD with WideResNet50. It can be seen that dimension reduction caused less impact to object categories than the texture categories. One example is \emph{carpet}, which is one of the texture categories and has largest dropped AUROC among all types. When reducing feature vectors to 128 dimensions, the image-level AUROC for \emph{carpet} dropped from 99.2\% to 98.0\%. Regarding the impact of batch size $n_b$, for object categories, the impact is negligible. For texture categories, setting $n_b$ to 16K or 32K actually outperformed $n_b=\text{all}$. 

Tables~\ref{table:training_time_wide_resnet50} and \ref{table:training_time_resnet18} list the training time of both WideResNet50 and ResNet18, respectively, on the category \emph{hazelnut}, which has the most training images among all MVTec AD categories. To benchmark the speed, we used only 1 CPU core to test, and pre-loaded all images into the memory beforehand to exclude the I/O time. When training on CPUs, we can see that the time was proportional to $k$, lengths of the reduced features, and the impact of batch size $n_b$ is small. This is because that the sampling of memory bank dominates the training. Regarding the inference time, which is not listed, we found that when using CPU alone, inference time is not linearly proportional to the dimension. This is because that forwarding an image throgh the backbone also requires computation time, especially on CPUs. Because dimension reduction cannot accelerate feature extraction, the total speedup is limited. 

To verify the benefit of dimension reduction, we tested our algorithm on \emph{Eyecandies}~\cite{:/conf/accv/2022/bonfiglioli_eyecandies}. This dataset has synthesized images of ten object types that were rendered under six lighting conditions. Each object type has 1000 training images, 100 validation images, 50 public testing images with annotations, and 400 private testing images without annotation. For each object type, we combined the training and validation images of all lighting conditions to form a large training set, which has 6600 images. Unlike MVTecAD, the images of eyecandies were resized into $256 \times 256$ pixels and tested without central cropping. Consequently, the image domain was divided into $32 \times 32$ patches. If each patch has a feature vector of 1024 dimensions, one image requires 4MB of space for $32 \times 32$ feature vectors in single precision, and consequently, 6600 training images require 25GB of memory, which is only affordable on high-end GPUs. By using our algorithm to incrementally reduce the dimensionality to 128, we can reduce the training time on GPUs to 3 hours. This demonstrates the efficiency of our dimension reduction algorithm. We also tested PaDiM~\cite{:/misc/arxiv/2020/defard_padim}, which is the state-of-the-art of this dataset. Since PaDiM is based on multivariate Gaussian distribution, we used only the first lighting condition (0) to train PaDiM. For fair comparison, we use the same 1024-dimensional feature vectors for both PadiM and PatchCore. We used testing image of lighting condition 0 to measure the AUROC. Table~\ref{table:eyecandies_auroc} lists AUROC on the public test set, which shows that with our dimension reduction algorithm, PatchCore can further boost the accuracy. We also tested PatchCore with images of lighting condition 0 only and reported the AUROC in Table~\ref{table:eyecandies_auroc}. It can be seen that using all lighting conditions indeed achieves better accuracy than using only one lighting condition, which consequently needs algorithms that can handle large training datasets like ours.

\begin{table}[tb!]
	\centering
	\caption{
Training time (seconds) on category \emph{hazelnut} of MVTec AD ~\cite{:/journal/ijvc/2021/bergmann_mvtec} with WideResNet50~\cite{:/conf/bmvc/2016/zagoruyko_wrn}. Please refer to the caption of Table~\ref{table:mvtec_ad_wide_resnet50} for the meaning of $k$ and $n_b$.
	}
	\label{table:training_time_wide_resnet50}
	\setlength\tabcolsep{3.5pt} 
		\begin{tabular}{lc|c|ccc|ccc}
			& $k$ 
			& N/A 
			& \multicolumn{3}{c|}{128} 
			& \multicolumn{3}{c}{256}  
			\\
			& $n_b$ 
			& N/A 
			& 16K & 32K & All 
			& 16K & 32K & All  
			\\
			\hline
			& GPU
			& 130  & 39  & 38  & 37  & 53  & 51  & 51  \\
			& CPU
			& 23998
			& 3333  & 2991  & 2965  & 5314  & 5073  & 4947  \\
		\end{tabular}		
	\end{table}

	\begin{table}[tb!]
		\centering
		\caption{
Training time (seconds)  on category \emph{hazelnut} of MVTec AD ~\cite{:/journal/ijvc/2021/bergmann_mvtec} with ResNet18~\cite{:/conf/cvpr/2016/he_resnet}. Please refer to the caption of Table~\ref{table:mvtec_ad_wide_resnet50} for the meaning of $k$ and $n_b$.
		}
		\label{table:training_time_resnet18}
		\setlength\tabcolsep{3.5pt} 
		\begin{tabular}{lc|c|ccc|ccc}
			& $k$ 
			& N/A 
			& \multicolumn{3}{c|}{128} 
			& \multicolumn{3}{c}{256}  
			\\
			& $n_b$ 
			& N/A 
			& 16K & 32K & All 
			& 16K & 32K & All  
			\\
			\hline
			& GPU
			& 128  & 37  & 36  & 35  & 50  & 49  & 48  \\
			& CPU
			& 26473
			& 2979  & 2975  & 2795  & 4952  & 4792  & 4841  \\
		\end{tabular}		
	\end{table}

\begin{table*}[tb!]
	\centering
	\caption{
AUROC on the dataset Eyecandies~\cite{:/conf/accv/2022/bonfiglioli_eyecandies} with PaDiM~\cite{:/misc/arxiv/2020/defard_padim} and ours, which used $k=128$ and $n_b=32$K. We used  WideResNet50~\cite{:/conf/bmvc/2016/zagoruyko_wrn}  as the feature extractor for both. The values were measured on the test images captured under lighting condition 0, whereas column L means lighting conditions of training images.
	}
	\label{table:eyecandies_auroc}
	\setlength\tabcolsep{3pt} 
	\begin{tabular}{lccc|cccccccccc|c}
& AUROC 
& Method
& L
& \makecell{Can.C.} 
& \makecell{Cho.C.} 
& \makecell{Cho.P.} 
& \makecell{Confet.} 
& \makecell{Gum.B.} 
& \makecell{Haz.T.} 
& \makecell{Lic.S.} 
& \makecell{Lollip.} 
& \makecell{Marsh.} 
& \makecell{Pep.C.} 
& \makecell{Avg.} \\
\hline
& \multirow[c]{3}{*}{Image}
& Ours & All
& 53.1 & 95.0 & 77.0 & 87.8 & 80.9 & 69.9 & 87.4 & 68.6 & 96.5 & 88.2 & 80.4 
\\
&
& Ours & 0
& 51.8 & 96.3 & 79.7 & 89.3 & 76.1 & 71.0 & 83.8 & 67.7 & 96.0 & 85.9 & 79.8 
\\
& 
& PaDiM & 0
& 47.2 & 93.6 & 71.2 & 87.7 & 86.5 & 69.9 & 86.2 & 67.3 & 96.2 & 84.5 & 79.0 
\\
\hline
& \multirow[c]{3}{*}{Pixel}
& Ours & All
& 91.6 & 97.3 & 86.6 & 98.2 & 93.5 & 85.7 & 94.1 & 96.3 & 97.0 & 97.3 & 93.8 
\\
&
& Ours & 0
& 89.8 & 97.0 & 86.2 & 98.2 & 93.6 & 86.1 & 93.6 & 96.4 & 96.6 & 96.3 & 93.4
\\
& 
& PaDiM & 0
& 92.7 & 96.6 & 87.1 & 97.6 & 93.9 & 88.1 & 93.7 & 96.8 & 97.5 & 97.0 & 94.1 
\\
	\end{tabular}		
\end{table*}



\section{Conclusion}

In this paper, we present an incremental dimension reduction algorithm to compress the memory bank of PatchCore. With our algorithm, we can reduce the dimensionality of features during the training with little overhead. To make PatchCore more practical, we would like further accelerate the algorithm in the future. One direction is eliminating redundant features before sampling, which is the bottleneck of training.


\bibliography{incremental_svd_for_patchcore}
\end{document}